\documentclass[sigconf, screen, nonacm]{acmart}
\usepackage{soul}
\setcopyright{none}
\newcommand{\mytablesize}{\small}
\newcommand{\mytableformat}{
  \setlength{\tabcolsep}{4pt}
  \renewcommand{\arraystretch}{0.9}
}
\renewcommand\footnotetextcopyrightpermission[1]{}
\AtBeginDocument{%
  }

\begin{document}

\title{Structure-Semantic Decoupled Modulation of Global Geospatial Embeddings for High-Resolution Remote Sensing Mapping}
\author{Jienan Lyu\textsuperscript{1}, Miao Yang\textsuperscript{2}, Jinchen Cai\textsuperscript{1}, Yiwen Hu\textsuperscript{1}, Guanyi Lu\textsuperscript{1}, Junhao Qiu\textsuperscript{1}, Runmin Dong\textsuperscript{1*}\\
\textsuperscript{1}Sun Yat-Sen University, \textsuperscript{2}Tsinghua University\\
\texttt{lvjn3@mail2.sysu.edu.cn}}

\authornote{Corresponding author: dongrm3@mail.sysu.edu.cn}



\begin{abstract}
Fine-grained high-resolution remote sensing mapping typically relies on localized visual features, which restricts cross-domain generalizability and often leads to fragmented predictions of large-scale land covers. While global geospatial foundation models offer powerful, generalizable representations, directly fusing their high-dimensional implicit embeddings with high-resolution visual features frequently triggers feature interference and spatial structure degradation due to a severe semantic-spatial gap. To overcome these limitations, we propose a Structure-Semantic Decoupled Modulation (SSDM) framework, which decouples global geospatial representations into two complementary cross-modal injection pathways. First, the structural prior modulation branch introduces the macroscopic receptive field priors from global representations into the self-attention modules of the high-resolution encoder. By guiding local feature extraction with holistic structural constraints, it effectively suppresses prediction fragmentation caused by high-frequency detail noise and excessive intra-class variance. Second, the global semantic injection branch explicitly aligns holistic context with the deep high-resolution feature space and directly supplements global semantics via cross-modal integration, thereby significantly enhancing the semantic consistency and category-level discrimination of complex land covers. Extensive experiments demonstrate that our method achieves state-of-the-art performance compared to existing cross-modal fusion approaches. By unleashing the potential of global embeddings, SSDM consistently improves high-resolution mapping accuracy across diverse scenarios, providing a universal and effective paradigm for integrating geospatial foundation models into high-resolution vision tasks. The code will be available at \url{https://github.com/jaco1b/SSDM-RS-SEG.}
\end{abstract}



\begin{teaserfigure}
  \centering
  \includegraphics[width=\textwidth]{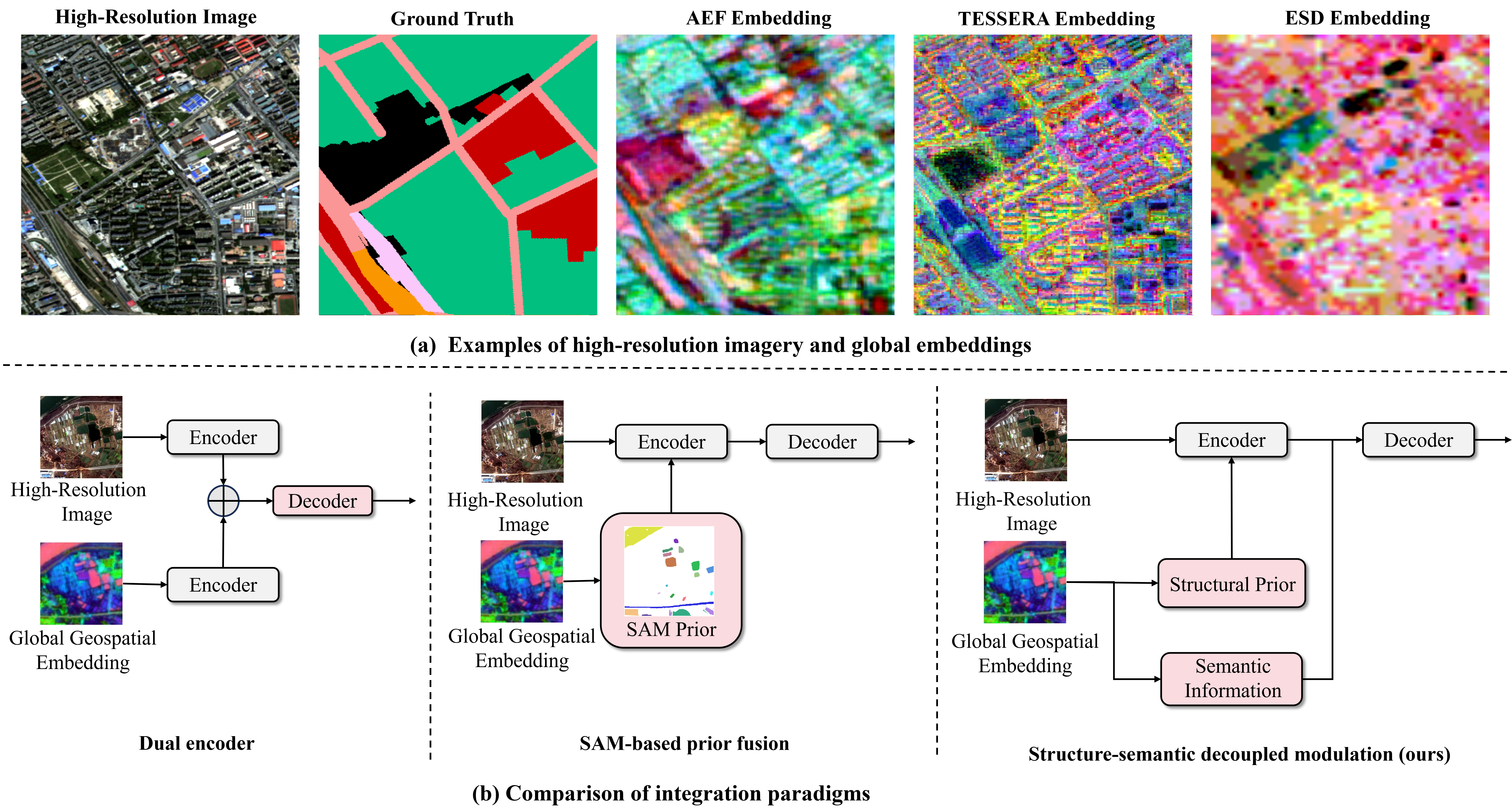}
  \caption{Comparison of different integration paradigms for global geospatial embeddings. (a) Visual examples of high-resolution imagery, ground-truth semantic mask, and diverse global geospatial embeddings. (b) Existing integration paradigms (e.g., dual-encoder and SAM-based fusion) alongside our proposed Structure-Semantic Decoupled Modulation (SSDM) approach.}
  \label{fig:teaser_motivation1}
  \vspace{12 pt}
\end{teaserfigure}

\maketitle
\thispagestyle{empty}
\pagestyle{empty}

\section{Introduction}
Recently, Geospatial Foundation Models (GFMs) have demonstrated remarkable capabilities in representing planetary-scale geographical patterns. Driven by self-supervised learning\cite{wang2022self} on massive Earth observation data\cite{fuller2023croma}, models such as AlphaEarth \cite{brown2025alphaearth}, ESD \cite{chen2026democratizing}, and TESSERA \cite{feng2025tessera} provide powerful and generalizable representations. However, the current application of these GFMs is primarily restricted to tasks at medium-to-low spatial resolutions, such as broad-scale vegetation and land cover mapping. While these models excel at capturing macroscopic geographical distributions, their coarse spatial and temporal resolutions prevent them from independently satisfying the fine-grained, dynamic demands of agricultural and urban applications. Crucially, this macroscopic observation capability is intrinsically complementary to the localized details of high-resolution remote sensing (RS) imagery. Therefore, shifting the paradigm from treating GFMs merely as standalone predictive models to leveraging their robust global representations as prior knowledge provides a principled approach to complement the localized focus of high-resolution mapping, constituting a significant yet underexplored direction in current research.

High-resolution remote sensing semantic segmentation models\cite{chen2018encoder} often rely on localized, high-frequency features\cite{audebert2018beyond}. While providing abundant texture details, this localized focus inherently limits cross-domain generalization. Moreover, when extracting large-scale continuous land cover categories, fine-grained texture and illumination fluctuations induce excessive intra-class variance, frequently leading to fragmented predictions. Conversely, in urban parcel-level mapping, this localized focus often results in rough and inaccurate boundaries\cite{li2024boundary}. Intuitively, the holistic structural priors and generalizable semantics encoded in global geospatial embeddings provide complementary information to supply the missing macroscopic context for fine-grained spatial features.

However, effectively integrating global embeddings into high-resolution architectures remains challenging. As illustrated in Fig.~1, a straightforward approach is to perform deep feature fusion\cite{ngiam2011multimodal} or treat global embeddings as an independent modality within a symmetric dual-encoder architecture\cite{hu2019acnet}. Yet, due to the substantial semantic-spatial disparity between high-dimensional implicit embeddings and fine-grained spatial features, such direct fusion often triggers feature interference and structure degradation. Alternatively, an intuitive solution to mitigate fragmentation is to leverage vision foundation models like the segment anything model (SAM)\cite{kirillov2023segment} to extract explicit structural priors. However, SAM is primarily optimized for object-centric natural images and exhibits suboptimal applicability to remote sensing data\cite{ren2024segment, zhang2024rs}, where land covers often lack clear boundaries and possess complex topologies. Furthermore, utilizing SAM to extract external structural priors introduces additional computational overhead. More critically, this independent extraction process overlooks the inherent spatial consistency of macroscopic representations, failing to fully exploit the advantages of global embeddings.

To overcome these limitations, we propose the structure-semantic decoupled modulation (SSDM) approach, a lightweight and effective paradigm for integrating global geospatial embeddings into high-resolution mapping. Instead of relying on computationally expensive external models, SSDM introduces two complementary cross-modal injection pathways tailored to different feature levels, thereby achieving a functional decoupling of structural and semantic modulation. Specifically, the structural prior modulation branch incorporates macroscopic spatial constraints as additive biases to directly modulate the self-attention maps within the high-resolution encoder, guiding fine-grained feature learning and mitigating prediction fragmentation. Subsequently, the global semantic injection branch integrates the complete global geospatial embeddings into the deepest layer of the encoder, providing global semantic compensation without disrupting the established spatial organization, thereby enhancing category-level discrimination and semantic coherence. 

The main contributions of this work are summarized as follows:
\begin{itemize}
\item We comprehensively explore the potential for integrating global geospatial embeddings to enhance high-resolution remote sensing mapping. The proposed SSDM approach effectively bridges the inherent semantic-spatial disparity between global representations and high-resolution features.
\item We design a lightweight and decoupled cross-modal modulation scheme comprising a structural prior branch and a global semantic branch, achieving robust fragmentation suppression and holistic structural consistency with minimal computational overhead.
\item Extensive experiments demonstrate that SSDM achieves state-of-the-art performance and exhibits consistent effectiveness across multiple global geospatial embedding databases, establishing a universal and scalable paradigm for adapting geospatial foundation models to high-resolution vision tasks.
\end{itemize}

\section{Related Work}

\subsection{High-Resolution Remote Sensing Semantic Segmentation}
High-resolution remote sensing semantic segmentation aims to achieve a fine-grained, pixel-level understanding of complex Earth surface scenes. Although existing Transformer architectures like Mask2Former have significantly enhanced context modeling\cite{cheng2022masked, cheng2021per}, high-resolution images are typically cropped into fixed-size patches during training and inference due to their extensive spatial coverage. Constrained by these physical cropping boundaries, models essentially extract visual features within a limited local receptive field\cite{chen2024integrating}. Consequently, high-resolution mapping frequently suffers from inter-class boundary confusion, limited cross-domain generalization, and fragmented predictions of large-scale land covers\cite{zhao2017pyramid, chen2018encoder, yuan2020object}.

To address these issues, recent studies have explored cross-domain generalization to alleviate distribution discrepancies by aligning the scene covariance between the source and target domains \cite{cao2024unsupervised, liu2022unsupervised}. In addressing large-scale mapping, RSET \cite{chen2025rest} captures broader contexts by expanding local windows, yet remain restricted by a finite visual receptive field. To mitigate boundary confusion and prediction fragmentation, some methods integrate foundation vision models like the Segment Anything Model (SAM)\cite{ravi2024sam,carion2025sam,kirillov2023segment} to provide zero-shot boundary priors\cite{wu2023samgeo}. However, their computationally intensive architectures incur substantial computational costs and often produce false boundaries in complex remote sensing scenes. Therefore, introducing global geospatial embeddings with macroscopic receptive fields as global priors emerges as a promising direction to overcome the limitations of high-resolution mapping.

\subsection{Geospatial Foundation Models}

Driven by large-scale Earth observation data, self-supervised geospatial representation learning\cite{sun2022ringmo, stewart2023ssl4eo} has yielded foundation models like AlphaEarth, ESD, and TESSERA. These models encode multi-source, multi-temporal\cite{cong2022satmae} observations into globally continuous embedding fields, transforming complex Earth observations into compact, highly generalizable high-dimensional vectors\cite{hong2024spectralgpt}.

Recently, the community has actively explored applying these global embeddings to downstream tasks\cite{lacoste2023geo, rolf2021generalizable}. For example, SatCLIP demonstrates their potential as effective auxiliary features for cross-regional image localization and zero-shot land cover mapping using simple shallow classifiers or multi-layer perceptrons \cite{klemmer2025satclip}.

Despite their broad potential in cross-domain applications, applying these high-dimensional global representations to high-resolution semantic segmentation remains a highly challenging and under explore. Existing paradigms primarily focus on coarse-grained predictions or rely on naive feature concatenation\cite{huo2025remote}. When applied to high-resolution mapping, such simple fusion fails to bridge the substantial semantic-spatial disparity between holistic implicit representations and fine-grained visual features. Naive concatenation often triggers semantic biases in deep network layers and disrupts the inherent spatial coherence of high-resolution imagery, thereby failing to fully utilize the advantages of global geospatial priors.

\subsection{Cross-Modal Fusion and Feature Modulation}

Incorporating multi-modal information is a standard approach to enhance representation capabilities. While early fusion methods relied on simple feature concatenation\cite{li2016lstm, ramachandram2017deep, hu2018squeeze}, recent advancements have introduced dynamic routing and Mixture of Experts (MoE) mechanisms\cite{shazeer2017outrageously, fedus2022switch}. However, these methods inevitably introduce substantial parameters and complex routing computations.

To circumvent these computational bottlenecks, recent studies in the fusion of RGB images with complementary modalities like depth (RGB-D) or thermal data (RGB-T) \cite{hazirbas2016fusenet, sun2019rtfnet} indicate that explicitly treating the auxiliary modality as a geometric or structural prior effectively improves representation learning. For example, depth information provides precise spatial layouts to distinguish targets with similar appearances but different shapes. Guided by these structure-aware fusion strategies, extensive works have achieved significant improvements in segmentation accuracy and boundary quality. However, unlike depth or thermal maps that provide explicitly aligned spatial geometries, global geospatial embeddings are inherently spatially implicit. Recognizing that this fundamental semantic-spatial disparity demands a paradigm shift beyond conventional structure-aware fusion, we propose the Structure-Semantic Decoupled Modulation (SSDM) framework to achieve the lightweight integration of global embeddings.

\begin{figure*}[t]
\centering
\includegraphics[width=\textwidth]{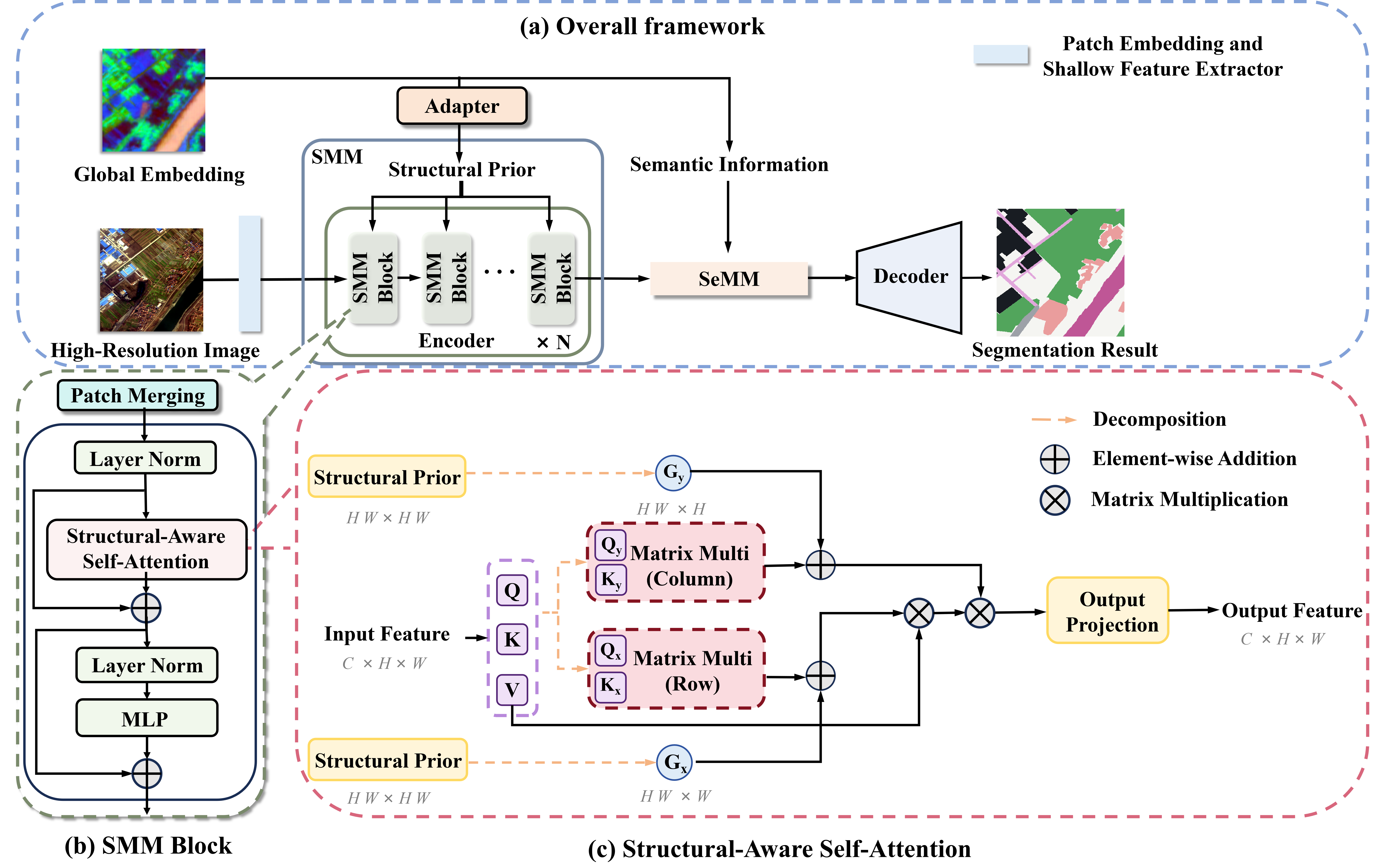}
\caption{
Overview of the proposed SSDM framework. (a) Overall framework. The adapted global embeddings are functionally decoupled to guide the encoder via SMM and complement semantic information via SeMM. (b) Architecture of the SMM block. (c) Structural-aware self-attention mechanism.
}
\label{fig:architecture}
\vspace{6pt}
\end{figure*}

\section{Methodology}
\subsection{Overall Framework}

To bridge the semantic–spatial gap between global geospatial embeddings and high-resolution visual features, we propose a structure–semantic decoupled modulation (SSDM) framework built upon Mask2Former, as illustrated in Fig.~\ref{fig:architecture} (a).

Given an input RGB image $\mathbf{I} \in \mathbb{R}^{3 \times H \times W}$ and its corresponding global geospatial embedding $\mathbf{E}$, we design two decoupled pathways to leverage the global embedding for complementary purposes. Specifically, for the structural modulation branch, $\mathbf{E}$ is first projected into a structure-aware representation via a lightweight adapter, which is used to derive structural priors for encoder-side modulation. In parallel, the semantic modulation branch directly utilizes the global embedding to extract high-level semantic information, without passing through the adapter, and injects it into the deep feature space before decoding. This design functionally decouples the structural and semantic modulation pathways, allowing each pathway to focus on its specific role without introducing cross-modal interference.

The framework consists of three components: a Mask2Former backbone, a Structural Modulation Module (SMM), and a Semantic Modulation Module (SeMM). The input image $\mathbf{I}$ is first processed by a patch embedding and shallow feature extractor. Subsequently, the backbone extracts multi-scale features $\{\mathbf{F}_l\}$ through the cascaded encoder blocks.

The SMM operates on the encoder side, where structural priors are derived from the adapted embedding via the structural branch and injected into each stage of the backbone in a stage-wise manner. This design enables structure-aware feature learning by introducing macroscopic spatial constraints into local feature extraction, thereby improving spatial consistency.

In parallel, the semantic branch bypasses the encoder and directly delivers global semantic information to the deep feature space. Specifically, after feature extraction, the pixel decoder produces mask features $\mathbf{M}$, and SeMM injects semantic priors into $\mathbf{M}$ through lightweight residual fusion, resulting in refined features $\mathbf{M}'$. The refined features are then fed into the transformer decoder for final prediction.

Overall, SSDM achieves a functional decoupling of structural and semantic modulation by introducing priors at different network stages: structural priors are injected early to guide spatial organization, while semantic priors are introduced late to enhance category-level consistency. This decoupled design enables complementary feature enhancement while effectively preventing cross-modal interference.

\subsection{Structural Modulation Module (SMM)}

High-resolution remote sensing imagery frequently exhibits pronounced intra-class variability and ambiguous boundaries due to complex textures and illumination variations, leading to fragmented predictions when relying solely on local visual features. To address this issue and inspired by \cite{yin2025dformerv2}, we introduce the Structural Modulation Module (SMM), which injects global structural priors into the encoder to guide feature learning with macroscopic spatial constraints. 

\noindent\textbf{Structural Prior Generation.}
Given the global geospatial embedding $\mathbf{E}$, we first derive a structure-aware representation:
\begin{equation}
\mathbf{G} = \phi_{\text{geo}}(\mathbf{E}),
\end{equation}
where $\phi_{\text{geo}}(\cdot)$ projects the embedding and computes pairwise interactions, yielding a dense affinity matrix $\mathbf{G} \in \mathbb{R}^{HW \times HW}$. This comprehensive relation matrix captures the macroscopic structural affinities and geometric distances between all pairs of spatial locations, serving as a global structural prior for subsequent relational modeling. To align with the decomposed attention mechanism, this full relation matrix $\mathbf{G}$ is subsequently sliced into direction-specific priors $\mathbf{G}_x \in \mathbb{R}^{HW \times W}$ and $\mathbf{G}_y \in \mathbb{R}^{HW \times H}$.

\noindent\textbf{Structure-Aware Attention.}
Given the backbone feature at stage $l$, denoted as $\mathbf{F}_l$, we first project it into a latent space:
\begin{equation}
\mathbf{X}_l = \mathbf{P}_l(\mathbf{F}_l),
\end{equation}
where $\mathbf{P}_l$ is a $1 \times 1$ convolution.

We then compute the query, key, and value projections:
\begin{equation}
\mathbf{Q}, \mathbf{K}, \mathbf{V} = \text{Proj}(\mathbf{X}_l).
\end{equation}

Unlike explicit geometric priors (e.g., depth maps) that can serve as hard attention masks, global geospatial embeddings provide implicit semantic-spatial representations rather than directly aligned spatial cues. Directly applying multiplicative gating would risk over-suppressing the fine-grained visual features extracted by the backbone. Therefore, instead of treating the structural prior as a hard mask, we incorporate it as a soft additive bias into the pre-softmax attention logits, allowing the model to modulate structural consistency while preserving local discriminative details.

To better model structured spatial patterns and improve efficiency, we follow established practices to adopt a decomposed attention formulation. Specifically, the attention is factorized into row-wise and column-wise components, where the queries and keys are correspondingly decoupled:
\begin{equation}
\mathbf{A}_x = \mathrm{Softmax}(\mathbf{Q}_x\mathbf{K}_x^\top + \mathbf{G}_x),
\end{equation}
\begin{equation}
\mathbf{A}_y = \mathrm{Softmax}(\mathbf{Q}_y\mathbf{K}_y^\top + \mathbf{G}_y),
\end{equation}
where the directional structural priors $\mathbf{G}_x \in \mathbb{R}^{HW \times W}$ and $\mathbf{G}_y \in \mathbb{R}^{HW \times H}$ are derived from the resized geometry map $\mathcal{G}_l$, which model horizontal and vertical spatial dependencies, respectively. In practice, the decomposed attention is applied sequentially to progressively build a global receptive field. The row-wise attention first modulates the value feature to capture horizontal contexts, $\mathbf{V}' = \mathbf{A}_x \mathbf{V}$, and the resulting intermediate representation is subsequently refined by the column-wise attention to capture complementary vertical dependencies, $\tilde{\mathbf{X}}_l = \mathbf{A}_y \mathbf{V}'$.

After the directional attention operations, the fused feature is passed through an output projection to aggregate the decomposed responses and map them back to the original feature space, followed by a simple multi-layer perceptron (MLP) to produce the final refinement term. This process effectively lifts the structural prior from a purely spatial feature to a relational space, enabling structure-aware modulation of attention.

\noindent\textbf{Residual Structural Injection.}
The final output feature is obtained via residual update:
\begin{equation}
\mathbf{F}_l' = \mathbf{F}_l + \Delta_l,
\end{equation}
where $\Delta_l$ is derived from the MLP output of $\tilde{\mathbf{X}}_l$. This residual formulation preserves the original feature distribution while progressively injecting structural priors.

\subsection{Semantic Modulation Module (SeMM)}

While the structural modulation module (SMM) improves spatial organization, accurate high-resolution mapping also requires reliable category-level semantics, especially for large homogeneous regions with similar local textures. We introduce a lightweight Semantic Modulation Module (SeMM), which injects global semantic context into the mask features after pixel decoding and before the transformer decoder.

\noindent\textbf{Design Motivation.}
Unlike structural information, which is used to guide multi-stage feature learning in the encoder, semantic information mainly serves as high-level contextual compensation. Injecting such semantic priors into early visual layers may interfere with the spatial structure formation of high-resolution features. Therefore, we introduce semantic modulation only at the head stage, after the pixel decoder has produced semantically richer mask features.

\noindent\textbf{Semantic Feature Encoding.}
Given the global geospatial embedding $\mathbf{E} \in \mathbb{R}^{64 \times H \times W}$, we first encode it into a compact semantic representation:
\begin{equation}
\mathbf{S} = \phi_{\text{sem}}(\mathbf{E}),
\end{equation}
where $\phi_{\text{sem}}(\cdot)$ is implemented as a lightweight convolutional encoder with two $3\times3$ convolutional layers and $\mathbf{S}$ denotes the encoded semantic feature. Different from the structural branch, this semantic branch does not share the structural projection, but uses an independent encoder to preserve semantic information for late-stage fusion.

\noindent\textbf{Late Semantic Injection.}
Let $\mathbf{M}$ denote the mask features produced by the pixel decoder. We first resize $\mathbf{S}$ to match the spatial resolution of $\mathbf{M}$:
\begin{equation}
\hat{\mathbf{S}} = \mathrm{Resize}(\mathbf{S}).
\end{equation}
We then concatenate the aligned semantic feature and the mask feature along the channel dimension, and use a lightweight projection function to generate a semantic compensation term:
\begin{equation}
\Delta_s = \phi_{\text{proj}}([\mathbf{M}, \hat{\mathbf{S}}]),
\end{equation}
where $[\cdot,\cdot]$ denotes channel concatenation and $\phi_{\text{proj}}(\cdot)$ is implemented as a lightweight $1\times1$ convolution.

Finally, the refined mask feature is obtained by residual update:
\begin{equation}
\mathbf{M}' = \mathbf{M} + \Delta_s.
\end{equation}

This residual formulation preserves the original mask feature representation while injecting global semantic context in a controlled manner. Unlike SMM, which modulates encoder-side multi-scale features for structural consistency, SeMM operates on the mask features after pixel decoding, thereby enhancing category-level consistency without interfering with spatial structure.

\subsection{Implementation Details}

Our framework is implemented in Detectron2 based on Mask2Former with a ResNet-50 backbone pretrained on ImageNet. Multi-scale features from res2–res5 are fed into the pixel and transformer decoders. All methods use a fixed 512×512 input and share identical training/validation splits.

For each RGB patch, we use aligned AEF features $\mathbf{E} \in \mathbb{R}^{64 \times H \times W}$, which are projected into compact representations and injected via two lightweight branches: a structural branch for encoder modulation and a semantic branch for residual fusion. All features are resized to match corresponding resolutions.

Training follows standard Mask2Former settings with AdamW optimizer, using a base learning rate of $1 \times 10^{-4}$ and weight decay of 0.05. Models are trained on 8 RTX 4090 GPUs with common data augmentations in Detectron2. All methods are trained under identical settings for fair comparison.

\begin{table*}[htbp]
\centering
\mytablesize
\mytableformat
\caption{Comparison results on the GID24 (4 m) and GID24 (2 m) subsets. All models are trained exclusively on the 4 m training set, and the 2 m test set is utilized to evaluate generalization. The best results are highlighted in bold.}
\label{tab:comparison}
\begin{tabular*}{\textwidth}{@{\extracolsep{\fill}}lcccccc@{}}
\toprule
Method & \multicolumn{3}{c}{GID24 (4 m)} & \multicolumn{3}{c}{GID24 (2 m)} \\
\cmidrule(lr){2-4} \cmidrule(lr){5-7}
       & OA (\%) & mIoU (\%) & mAcc (\%) & OA (\%) & mIoU (\%) & mAcc(\%) \\
\midrule
Baseline         & 75.29 & 39.08 & 48.87 & 72.29 & 34.65 & 44.91 \\
Dual-Encoder \cite{jia2024geminifusion}         & 80.41 & 42.28 & 53.70 & 76.92 & 38.31 & 49.11 \\
SAM-based     \cite{ma2024sam}       & 81.95 & 44.12 & 55.83 & 78.76 & 40.27 & 51.34  \\
DFormerv2$^{*}$ \cite{yin2025dformerv2}     & 84.24 & 48.87 & 60.73 & 81.32 & 45.50 & 57.21 \\
\textbf{SSDM (Ours)} & \textbf{85.13} & \textbf{50.01} & \textbf{62.25} & \textbf{82.64} & \textbf{47.32} & \textbf{59.08} \\
\bottomrule
\end{tabular*}
\vspace{10 pt}
{\noindent\textsuperscript{*}\parbox[t]{0.9\linewidth}{DFormerv2 based on the official implementation, with an added AEF adapter to enable AEF input.}}
\vspace{10 pt}
\end{table*}

\begin{table}[t]
\centering
\mytablesize
\mytableformat
\caption{Per-class performance on the GID24 (4 m) subset.}
\label{tab:per_class}
\begin{tabular}{@{}p{0.34\columnwidth}cccc@{}}
\toprule
Class & \multicolumn{2}{c}{Baseline} & \multicolumn{2}{c}{SSDM (Ours)} \\
\cmidrule(lr){2-3} \cmidrule(lr){4-5}
 & Acc (\%) & IoU (\%) & Acc (\%) & IoU (\%) \\
\midrule
Industrial land   & 77.14 & 58.12 & 87.21 & 74.35 \\
Paddy field       & 75.57 & 57.86 & 85.43 & 74.02 \\
Irrigated field   & 84.82 & 70.59 & 95.89 & 90.31 \\
Dry cropland      & 71.93 & 53.68 & 81.32 & 68.67 \\
Garden land       & 44.11 & 24.64 & 49.87 & 31.52 \\
Arbor forest      & 77.45 & 66.67 & 87.56 & 85.29 \\
Shrub forest      & 45.80 & 14.40 & 51.78 & 18.43 \\
Park              & 34.70 & 28.12 & 39.23 & 35.97 \\
Natural meadow    & 83.61 & 67.43 & 94.52 & 86.27 \\
Artificial meadow & 53.52 & 36.64 & 60.51 & 46.87 \\
River             & 56.72 & 47.38 & 64.12 & 60.61 \\
Urban residential & 80.70 & 63.34 & 91.23 & 81.03 \\
Lake              & 81.78 & 59.52 & 92.45 & 76.15 \\
Pond              & 51.63 & 23.86 & 58.37 & 30.52 \\
Fish pond         & 76.96 & 57.86 & 87.01 & 74.03 \\
Snow              & 22.02 & 12.78 & 24.89 & 16.35 \\
Bareland          & 81.26 & 61.53 & 91.87 & 78.72 \\
Rural residential & 73.32 & 59.23 & 82.89 & 75.78 \\
Stadium           & 59.99 & 40.70 & 67.82 & 52.07 \\
Square            & 41.41 & 19.28 & 46.81 & 24.66 \\
Road              & 77.06 & 57.85 & 87.12 & 74.01 \\
Overpass          & 72.03 & 55.53 & 81.43 & 71.04 \\
Railway station   & 62.02 & 34.10 & 70.12 & 43.63 \\
Airport           & 61.06 & 46.88 & 69.03 & 59.98 \\
\midrule
\textbf{Overall} & \multicolumn{2}{c}{75.29 / 39.08} & \multicolumn{2}{c}{\textbf{85.13 / 50.01}} \\
\bottomrule
\end{tabular}
\vspace{10 pt}
\end{table}

\begin{figure*}[t]
\centering
\includegraphics[width=\textwidth]{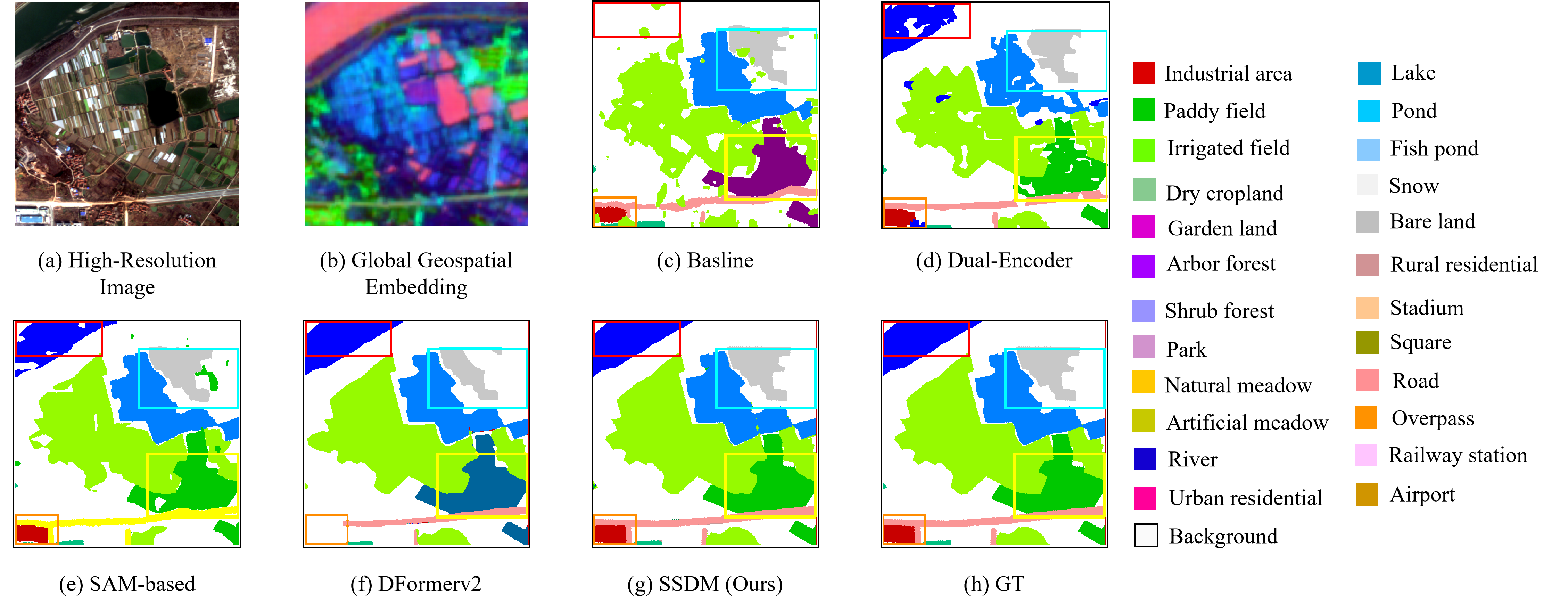}
\caption{Qualitative comparison of segmentation results. The colored bounding boxes highlight complex regions where existing methods suffer from severe fragmentation or boundary inconsistency, whereas our SSDM maintains superior structural coherence and precise delineation.}
\label{fig:qualitative}
\vspace{10 pt}
\end{figure*}

\section{Experiment}
\subsection{Datasets and Evaluation}
\textbf{Datasets.}
GID24 \cite{liu2024global} is a high-resolution remote sensing dataset designed for large-scale land-cover mapping with a spatial resolution of 4 m. 
The dataset contains 24 semantic categories, including commonly used land cover categories (e.g., paddy field, irrigated field, dry cropland, arbor
forest, shrub forest, natural meadow, artificial meadow, river, lake, and bare land) and fine-grained land use categories (e.g., industrial area, garden land, park, urban residential, rural residential,
stadium, square, road, overpass, railway station, and airport). 

Following the standard data processing, all images are cropped into $512 \times 512$ patches and we use RGB bands in the experiments. We sample representative subsets from the original splits to conduct evaluations. Specifically, for comparison and ablation studies, we construct a subset of 4,000 patches, comprising 3,000 for training and 1,000 for testing. 
At the same time, we introduce an additional out-of-domain test set to evaluate the model robustness. This dataset comprises 2-meter resolution imagery collected from the GaoFen and ZiYuan satellite series, including the identical 24 semantic categories as the training set.

For the fusion experiments, we extract and spatially align the corresponding global geospatial embeddings for each sample region. Our primary experiments are conducted using the AEF embeddings \cite{brown2025alphaearth} at a 10 m spatial resolution. It is noteworthy that a temporal misalignment exists in our multimodal setup. While the high-resolution imagery and ground-truth labels were acquired in 2015 and 2016, the earliest available global geospatial embeddings (e.g., AEF annual composites) date back to 2017. Consequently, we pair the 2015, 2016 images with the temporally closest 2017 embeddings. Although this inevitable temporal gap introduces certain ground information discrepancies (e.g., land-cover changes over time), it constitutes a ubiquitous and realistic challenge\cite{claverie2018harmonized} in multi-source remote sensing fusion. Therefore, evaluating the models under this stringent, temporally unaligned setting can reflect their robustness and noise tolerance capability. To further verify the generalizability of our approach, we also incorporate two other types of representations, including TESSERA embeddings \cite{feng2025tessera} at a 10 m resolution and ESD embeddings \cite{chen2026democratizing} at a 30 m resolution. 

\textbf{Evaluation metrics.}
Semantic segmentation performance is evaluated using Overall Accuracy (OA), Mean Intersection over Union (mIoU), and Mean Accuracy (mAcc). 
In our implementation, these metrics are computed using the semantic segmentation evaluator in Detectron2 \cite{wu2019detectron2}. Specifically, OA measures the pixel-wise accuracy over all valid pixels, mIoU computes the average intersection-over-union across all semantic classes, and mAcc reports the average class-wise pixel accuracy. Among these metrics, mIoU is adopted as the primary criterion for quantitative comparison.

\subsection{Comparison Results}

We evaluate all methods on the GID24 subsets at two spatial resolutions (4 m and 2 m). All models are trained exclusively on the 4 m training set. The 4 m test set is utilized for in-domain evaluation, while the 2 m test set is directly applied to assess the cross-resolution generalizability of the models.

As shown in Table~\ref{tab:comparison}, compared to the HR-only baseline, all methods incorporating global geospatial embeddings achieve improved segmentation accuracy. The dual-encoder model obtains steady gains; the SAM-based variant and DFormerv2$^{*}$ further improve the results by leveraging explicit structural priors. In comparison, the proposed SSDM effectively avoids feature interference caused by direct multimodal fusion by explicitly decoupling structural priors and semantic information, thereby achieving the highest OA, mIoU, and mAcc across both test subsets.

Under the cross-resolution evaluation (4 m $\rightarrow$ 2 m), all methods experience performance degradation due to spatial distribution shifts. However, SSDM maintains the highest accuracy among all compared methods under this setting. This indicates that the decoupled modulation architecture can effectively transfer scale-invariant structural and semantic priors, mitigating the feature misalignment caused by resolution discrepancies.

Table~\ref{tab:per_class} reports the per-class performance on the 4 m subset, demonstrating that SSDM achieves IoU improvements across all 24 land-cover categories compared to the baseline. The most significant gains are concentrated in large-scale continuous land covers, such as irrigated fields (+19.72\%) and natural meadows (+18.84\%), as well as land-use categories with distinct distribution patterns, including urban residential (+17.69\%) and industrial land (+16.23\%). These substantial improvements are attributed to the macroscopic spatial coherence constraints introduced by the global priors, which enable the structural modulation branch to effectively suppress prediction fragmentation within homogeneous regions.

Conversely, categories exhibiting relatively marginal improvements include snow (+3.57\%), shrub forests (+4.03\%), squares (+5.38\%), and ponds (+6.66\%). These classes typically suffer from strong boundary ambiguity, smaller spatial scales, or extremely high inter-class similarity, such as the frequent confusion between shrub and arbor forests. This performance variance suggests that while global-scale embeddings excel at providing macroscopic context, their yielded gains are relatively limited when the task heavily relies on distinguishing highly fine-grained local features.

To further validate the segmentation outputs, Fig.~\ref{fig:qualitative} presents qualitative comparisons across the evaluated models. When processing large-scale homogeneous regions, SSDM generates predictions with higher spatial coherence, effectively reducing fragmentation and enhancing semantic consistency. In contrast, the baseline and dual-encoder models are prone to inconsistent predictions across adjacent regions due to their over-reliance on localized visual cues. Furthermore, the SAM-based approach frequently suffers from over-segmentation, as it is easily misled by inaccurate external boundary priors when handling the complex topologies of remote sensing land covers. Overall, SSDM consistently yields more accurately aligned boundaries, confirming the substantial effectiveness of the decoupled design in bridging the semantic-spatial gap between high-resolution features and global macroscopic priors.

\subsection{Ablation Study}

We conduct ablation experiments on the GID24 subset to evaluate the effectiveness of different integration strategies for global geospatial embeddings. All variants are trained under identical settings for fair comparison.

As shown in Table~\ref{tab:ablation}, both semantic modulation and structural modulation improve performance over the baseline, indicating that global embeddings provide useful complementary information.

\begin{table}[t]
\centering
\mytablesize
\setlength{\tabcolsep}{8pt}
\renewcommand{\arraystretch}{1.2}
\caption{Ablation study of the structural (SMM) and semantic (SeMM) modulation modules on the GID24 (4m) subset.}
\label{tab:ablation}
\begin{tabular*}{\columnwidth}{@{\extracolsep{\fill}}lccccc@{}}
\toprule
Method & SMM & SeMM & mIoU (\%) & mAcc (\%) & OA (\%) \\
\midrule
Baseline           & $\times$ & $\times$ & 39.08 & 52.87 & 78.31 \\
\shortstack[l]{+Semantic\\Modulation}  & $\times$ & \checkmark & 43.21 & 56.02 & 80.15 \\
\shortstack[l]{+Structure\\Modulation} & \checkmark & $\times$ & 45.89 & 58.74 & 81.63 \\
\textbf{Full} & \checkmark & \checkmark & \textbf{50.01} & \textbf{62.35} & \textbf{83.92} \\
\bottomrule
\end{tabular*}
\end{table}

Specifically, the SeMM yields quantitative gains, demonstrating that injecting global semantic context helps stabilize category-level predictions. However, the improvement is limited, as it does not explicitly address spatial structure. In the meanwhile, the SMM achieves larger performance gains, highlighting the importance of incorporating structure-aware priors into the feature learning process. By injecting structural information into the encoder in a stage-wise manner, the model produces more spatially coherent predictions and reduces fragmentation.

The full model consistently achieves the best performance across all metrics. This confirms that structural and semantic information play complementary roles: structural modulation enhances spatial consistency, while semantic modulation improves category-level discrimination. Crucially, this progressive improvement confirms that the performance gains are from the decoupled modulation architecture rather than the mere introduction of additional global embeddings.

\subsection{Generalization Across Global Embeddings}

We evaluate the generalization capability of SSDM by directly applying a model trained with AEF to other global embedding sources, including TESSERA and ESD, under a zero-shot cross-embedding transfer setting (i.e., without any parameter fine-tuning).

As shown in Table~\ref{tab:embedding_generalization}, all embedding variants consistently outperform the RGB-only baseline, demonstrating the effectiveness of incorporating global priors. Notably, SSDM maintains stable performance across different embedding types, despite substantial differences in spatial resolution and representation structure.

\begin{table}[t]
\centering
\mytablesize
\mytableformat
\caption{Generalization performance across heterogeneous global geospatial embeddings.}
\label{tab:embedding_generalization}
\begin{tabular}{@{}p{0.42\columnwidth}ccc@{}}
\toprule
Setting & OA (\%) & mIoU (\%) & mAcc (\%) \\
\midrule
Without embedding (HR only) & 75.29 & 39.08 & 48.87 \\
\midrule
With AEF      & 85.13 & 50.01  & 62.25 \\
With TESSERA  & 84.57 & 49.38 & 61.30 \\
With ESD      & 83.29 & 48.21 & 59.61 \\
\bottomrule
\end{tabular}
\vspace{8pt}
\end{table}

These results indicate that SSDM is embedding-agnostic and does not rely on a specific embedding distribution. The robustness arises from the decoupled design: structural modulation focuses on spatial consistency, while semantic modulation provides controlled global context, enabling stable adaptation across heterogeneous embeddings.



\subsection{Efficiency Analysis}

We compare both segmentation accuracy and computational cost on the GID24 (4 m) subset, as summarized in Table~\ref{tab:complexity}.

\begin{table}[t]
\centering
\mytablesize
\mytableformat
\caption{Computational cost and efficiency analysis on the GID24 (4 m) subset.}
\label{tab:complexity}
\begin{tabular*}{\columnwidth}{@{\extracolsep{\fill}}lccccc@{}}
\toprule
Method & mIoU (\%) & \shortstack{Par.\\(M)} & \shortstack{FLOPs\\(G)} & \shortstack{Lat.\\(ms)} & \shortstack{Mem.\\(GB)} \\
\midrule
Baseline & 39.08 & 44.00 & 66.95 & 20.51 & 0.43 \\
Dual-Encoder & 42.28 & 59.05 & 172.43 & 39.00 & 1.04 \\
SAM-based & 44.12 & 260.08 & 388.26 & 60.12 & 1.89 \\
DFormerv2$^{*}$ & 48.87 & 26.72 & 36.01 & 39.16 & 0.40 \\
SSDM (Ours) & 50.01 & 49.68 & 195.20 & 33.74 & 1.07 \\
\bottomrule
\end{tabular*}

\vspace{4 pt}
\parbox{\columnwidth}{\raggedright
\textbf{Note.} Par.: parameters (M). Lat.: average inference time per $512\times512$ image in milliseconds. Mem.: peak memory (GB). 
DFormerv2$^{*}$ denotes the official DFormerv2 with an added AEF adapter. 
For SAM-based methods, 260.08M counts total parameters, while trainable parameters are 47.38M (SAM encoder frozen).}
\end{table}
\vspace{7pt}

SSDM achieves the best segmentation performance while maintaining a favorable accuracy-efficiency trade-off. Compared with the RGB-only baseline, SSDM significantly improves mIoU with only a moderate increase in parameters, indicating that the performance gain is not simply due to model scaling. 

Compared with the dual-encoder design, SSDM achieves higher accuracy with fewer parameters and lower latency, avoiding redundant feature extraction. Although SSDM introduces slightly higher FLOPs, its lower latency suggests better practical efficiency. Compared with the SAM-based approach, SSDM achieves higher accuracy with substantially reduced computational overhead, avoiding the heavy cost of external foundation models.

Overall, SSDM provides a strong balance between performance and efficiency, making it suitable for large-scale high-resolution remote sensing applications.

\subsection{Discussion and Limitations}

The experimental results demonstrate that SSDM improves both spatial coherence and semantic consistency by functionally decoupling structural and semantic information. Structural modulation enhances global spatial organization and reduces fragmentation, particularly for large homogeneous regions, while semantic modulation improves category-level discrimination and stabilizes predictions. Detailed per-class improvements in Table~\ref{tab:per_class} further validate this observation.

Despite these improvements, several limitations remain. SSDM still struggles with categories characterized by high intra-class variability, ambiguous boundaries, or small object scales, such as shrub forest, snow, and square, as reflected in the per-class results in Table~\ref{tab:per_class}. These remaining challenges typically are from highly complex spatial distributions and extremely subtle inter-class differences.

Moreover, while structural modulation improves global consistency, it may introduce oversmoothing effects on fine-grained details, leading to reduced accuracy on small or irregular structures. The semantic branch, although effective in stabilizing predictions, remains limited in capturing subtle distinctions in highly ambiguous regions.

Future work will explore more adaptive semantic modeling for fine-grained discrimination, as well as detail-preserving structural constraints to alleviate oversmoothing. Extending the framework to more challenging tasks such as instance segmentation and change detection is another promising direction.

\section{Conclusion}
In this paper, we propose SSDM, a lightweight structure-semantic decoupled modulation approach for integrating global geospatial embeddings into high-resolution remote sensing mapping. The proposed approach addresses the semantic-spatial gap between high-dimensional global representations and fine-grained visual features by achieving a functional decoupling of structural and semantic modulation pathways. Specifically, a structural prior modulation branch guides the encoder-level feature extraction with holistic spatial constraints, while the global semantic injection branch enhances category-level discrimination at the encoder-decoder interface. Extensive experiments demonstrate the effectiveness of our approach, indicating that SSDM offers an efficient solution for enhancing high-resolution vision tasks with global geospatial embeddings. Moreover, comprehensive ablation studies and visualizations further validate the advantages of our dual-path design over conventional unified embedding fusion strategies, highlighting its potential for broader deployment in practical remote sensing monitoring systems.

\clearpage
\bibliographystyle{ACM-Reference-Format}
\bibliography{refs}

\end{document}